\documentclass[conference]{IEEEtran}
\IEEEoverridecommandlockouts
\usepackage{cite}
\usepackage{amsmath,amssymb,amsfonts}

\usepackage{siunitx}
\usepackage{algorithm}
\usepackage{algpseudocode}
\usepackage{graphicx}
\usepackage{subcaption}
\usepackage{textcomp}
\usepackage{multirow}
\usepackage{float}
\usepackage{xcolor}
\usepackage{tabularx}
\usepackage{booktabs}
\usepackage{multirow}

\newcolumntype{Y}{>{\centering\arraybackslash}X}
\begin{document}

% \title{Uncertainty-Aware Imitation Learning For Chemistry Lab Automation\\

\title{SAFE-CHEM: Uncertainty-Aware Policy Switching for Robust Robotic Chemistry}

\author{Laura Jones$^{1}$, Shazil Shahzad$^{1}$, Ayesha Sana$^{1}$ and Gabriella Pizzuto$^{1, 2}$\\
\thanks{$^{1}$ School of Computer Science \& Informatics, University of Liverpool, UK. Corresponding author email: \texttt{Laura.Jones@liverpool.ac.uk}}%
\thanks{$^{2}$ Department of Chemistry, University of Liverpool, UK.} %
 \thanks{This work was supported by the Royal Academy of Engineering under the Research Fellowship Scheme and the Google DeepMind Research Ready Scheme, and EPSRC through the Digital and Automated Materials Chemistry Centre for Doctoral Training (EP/Y03502X/1).}}

% \author{\IEEEauthorblockN{\blackout{Laura Jones}}
% \IEEEauthorblockA{\textit{School of Computer Science \& Informatics} \\
% \textit{University of Liverpool}\\
% Liverpool, UK \\
% Laura.Jones@liverpool.ac.uk}
% \and
% \IEEEauthorblockN{2\textsuperscript{nd} Shazil Shahzad}
% \IEEEauthorblockA{\textit{School of Computer Science \& Informatics} \\
% \textit{University of Liverpool}\\
% Liverpool, UK \\
% }
% \and
% \IEEEauthorblockN{3\textsuperscript{rd} Ayesha Sana}
% \IEEEauthorblockA{\textit{School of Computer Science \& Informatics} \\
% \textit{University of Liverpool}\\
% Liverpool, UK \\
% }
% \and
% \IEEEauthorblockN{4\textsuperscript{th} Gabriella Pizzuto}
% \IEEEauthorblockA{\textit{School of Computer Science \& Informatics} \\
% \textit{University of Liverpool}\\
% Liverpool, UK \\
% }

% 
% \and
% \and

\maketitle

\begin{abstract}
The deployment of autonomous robotic systems in chemistry laboratories is accelerating experimental workflows and providing the foundational data for AI-driven scientific discovery.
However, despite the success of data-driven methods in acquiring dexterous skills, safety remains a primary barrier to their deployment in high-risk domains, such as early-stage materials chemistry experiments.
Specifically, learning-based policies frequently struggle to distinguish between safe and unsafe actions, leading to overconfident extrapolation and potentially catastrophic failures.
To mitigate these safety risks, we propose \textbf{SAFE-CHEM}, an uncertainty-aware framework designed for robust, learning-based robotic chemists.
Our approach leverages an ensemble of recurrent neural network-based imitation learning policies to quantify epistemic uncertainty online through the variance of action predictions. 
By characterising the success-conditioned density of this variance using kernel density estimation, we introduce a hybrid control architecture that autonomously switches from the learned policy to a deterministic, rule-based backup controller when uncertainty exceeds a calibrated safety threshold.
We evaluate \textbf{SAFE-CHEM} across three fundamental laboratory manipulation tasks, where our empirical results demonstrate that this hybrid strategy improves overall task success rates and reduces critical safety violations compared to traditional single-policy baselines. 
Finally, we demonstrate the practical viability of the framework through zero-shot sim-to-real transfer onto a physical Franka Production 3 robot manipulator.

% While data-driven methods, such as imitation learning, enable robots to acquire dexterous manipulation skills from expert demonstrations, their deployment remains restricted by critical safety concerns. 
% Specifically, vanilla behavioural cloning policies often struggle to distinguish between safe and unsafe actions, leading to overconfident extrapolation and catastrophic, unsafe failures during high-risk procedures. 
\end{abstract}

% \begin{IEEEkeywords}
% component, formatting, style, styling, insert
% \end{IEEEkeywords}

\section{Introduction}
\label{section:intro}
% [Intro paragraph on the benefits of chem lab automation] 
The integration of robotics and artificial intelligence has the potential to accelerate traditional human-driven chemistry experiments towards autonomous and more efficient workflows~\cite{Cooper2025, Tom2024}. 
The benefits of robotic chemists have already been demonstrated globally across diverse workflows~\cite{Yoshikawa2025, Li2025, burger2020}.
These systems can execute a broad spectrum of experiments by utilising general-purpose robots, but most remain restricted to rigid, non-reconfigurable workspaces where changing to novel experiments often incurs substantial downtime~\cite{xie2023}.

% Paradoxically, while robotic manipulators introduce new operational risks, they simultaneously serve as vital instruments for human safety by insulating researchers from hazardous environments and high-risk procedures, exemplified by tangential applications such as nuclear decommissioning~\cite{tu2024}. 

% [paragraph on learning-based methods for chem lab automation] 
\par To overcome the limitations of rigid traditional methods in chemistry lab automation, recent works in robotic manipulation for chemistry have pivoted towards learning-based methods, specifically using imitation learning (IL) and reinforcement learning (RL)~\cite{Tom2024}. 
Unlike traditional automation that relies on precise coordinate-based trajectories, learning-based approaches allow manipulators to acquire complex, dexterous skills \textit{e.g.}, powder weighing~\cite{radulov2026}, pouring of fluids~\cite{Darvish2025}, or sample scraping~\cite{pizzuto2024}. 
Although IL offers a sample-efficient alternative to RL, its application to chemistry remains under-explored.
In chemistry, the difficulty of hand-engineering reward functions for multi-step procedures often makes expert demonstrations a more practical alternative.

\begin{figure}[]
    \centering
    \includegraphics[width=0.45\textwidth]{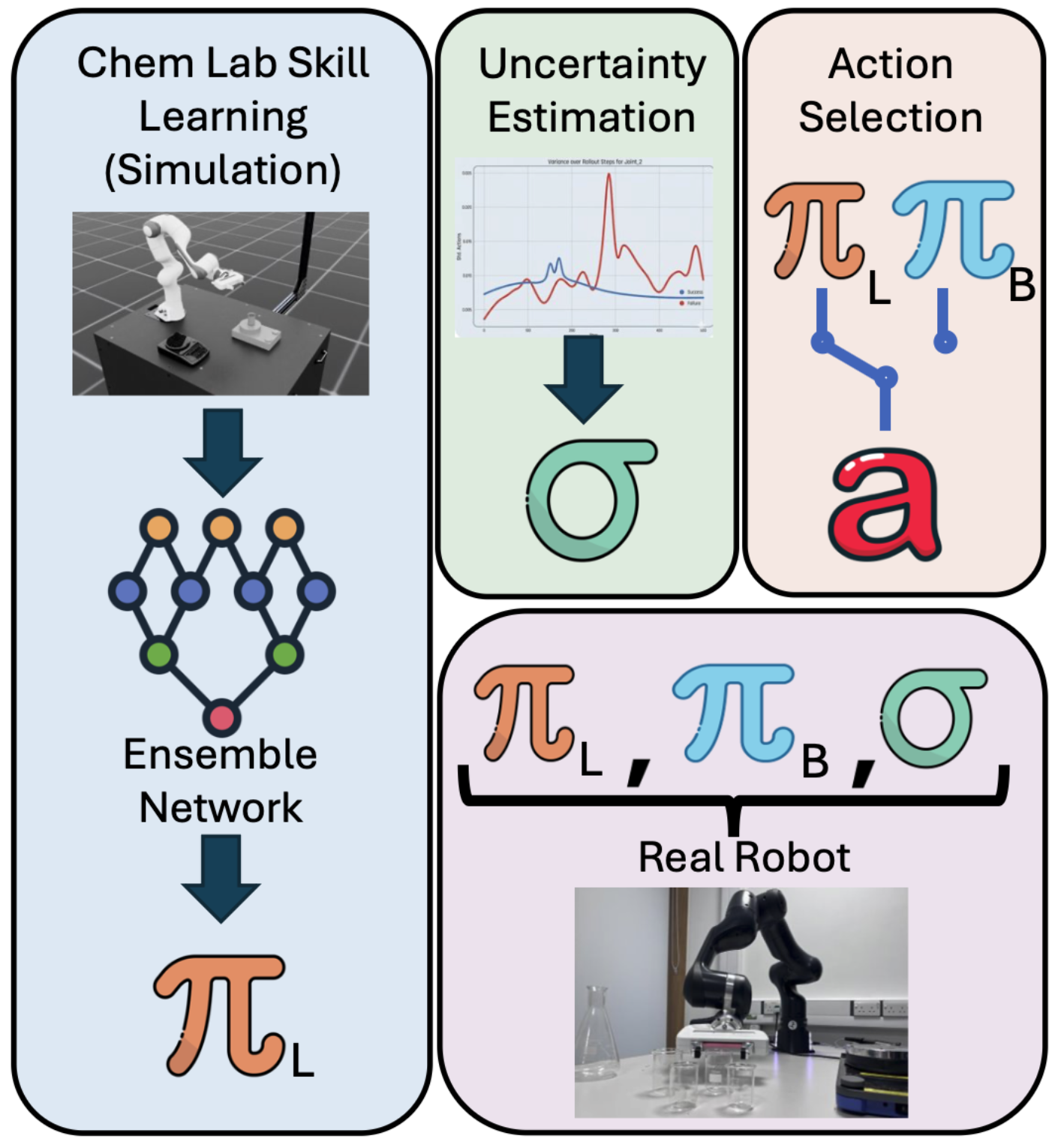}
    \caption{An overview of our method \textbf{SAFE-CHEM}, which consists of four stages: (1) training an ensemble-based imitation learning policy ($\pi_{L}$) in simulation for robotic chemistry manipulation tasks; (2) estimating the epistemic uncertainty $\sigma$ through ensemble variance; (3) action selection that monitors $\sigma$ to gate control between $\pi_{L}$ and a fail-safe rule-based backup robot controller ($\pi_{B}$); and (4) real-world demonstration on a robotic system using zero-shot task transfer.}
    \label{fig:intro_image}
\end{figure}

\par Despite the flexibility offered by these learning-based methods, the widespread adoption of robotic systems in human-centric chemistry laboratories remains contingent on overcoming critical safety hurdles~\cite{Leong2024}, \textit{e.g.}, chemical spillage~\cite{veeramani2025} and reactive collision avoidance in constrained environments~\cite{Longley2026}. 
Beyond protecting human scientists from hazardous environments and high-risk procedures, autonomous systems alleviate the burden of repetitive, time-intensive tasks, thereby reallocating human cognitive resources towards complex problem-solving~\cite{Cooper2025}.
Addressing the safety challenges within these tasks, from robust robotic manipulation to intelligent decision-making by autonomous agents, is critical for the reliable deployment of these systems~\cite{Leong2024}.

% [why imitation learning alone is not good enough in chem labs i.e., you cannot have an overconfident policy and hence motivate the need to handle uncertainties and introduce a switching behaviour when confidence is low] 
While IL has emerged as a prominent paradigm for training these autonomous agents, its reliability in practice remains fundamentally bounded by the density and diversity of the provided expert demonstrations.
A critical challenge in chemistry laboratory-based IL is the distributional shift that occurs when an agent encounters states outside the expert manifold \textit{e.g.}, novel equipment or unexpected reaction dynamics.
Because vanilla IL models often fail to distinguish between epistemic uncertainty (lack of data) and aleatoric uncertainty (inherent noise), they are prone to overconfident extrapolation.
In a safety-critical environment, as is a chemistry lab, a policy may attempt to execute an action in an out-of-distribution (OOD) state~\cite{xu2025F}, potentially leading to catastrophic failures~\cite{Leong2024} without any internal mechanism to trigger a safety intervention or seek human input. 
% Consequently, for robotic systems to be truly viable in chemistry labs, they need to have a risk-aware mechanism that distinguishes between tasks it can complete with high certainty and those that exceed a predefined uncertainty threshold, necessitating an immediate transition to a fail-safe state.

To address this, we propose a modular framework \textbf{SAFE-CHEM} that treats failure prevention in chemistry lab manipulation tasks as a real-time uncertainty-quantification problem, as illustrated in Fig.~\ref{fig:intro_image}.
Our method utilises an ensemble of neural network policies, whereby it monitors epistemic drift during inference to trigger a switching mechanism that transitions control back to a fail-safe state whenever the model's confidence falls below a critical threshold, ensuring that the robot's adaptive behaviour does not come at the cost of laboratory safety.

In summary the contributions of this work are:

(1) An ensemble-based imitation learning architecture specifically tailored for chemistry laboratory robotic manipulation, which enhances task robustness by capturing distribution of uncertainty across task rollouts to provide a more resilient policy than traditional single-model approaches;

(2) An uncertainty-aware, policy-agnostic safety framework that monitors variance across ensemble predictions to autonomously identify high-risk states and trigger a switching mechanism, effectively balancing task success with the frequency of autonomous intervention;

(3) An empirical quantification of the relationship between policy confidence and task success, providing a systematic analysis of how safety thresholds can be tuned based on the specific risk profile of a laboratory task; 

(4) A zero-shot real-world demonstration of our framework onto a physical robotic manipulator.

\section{Related Work}
\label{sec:related_work}

\subsection{Robotics for Laboratory Automation}
The adoption of robotic systems in chemistry labs is slowly evolving towards adaptive, modular systems capable of learning new laboratory skills.
Current approaches for developing these low-level behaviours typically fall into two categories: RL~\cite{radulov2026, kadokawa2023, pizzuto2024} or more recently, IL.
For instance, Xu et al.~\cite{Xu2025R} demonstrated the first application of imitation learning in a materials laboratory by training a a behaviour cloning with a recurrent neural network (BC-RNN) policy on 50 kinesthetic demonstrations to achieve autonomous cathode painting on fuel cell pellets. 
More recently, Suzuki et al.~\cite{suzuki2026} proposed TVF-DiT, a compact imitation learning framework that aligns vision and vision-language foundation models with a diffusion transformer-based action expert to enable efficient, high-success-rate robotic laboratory automation on low-VRAM hardware.
Despite the promise of these learning-based skills, their deployment in safety-critical chemistry labs is hindered by a lack of failure-prediction mechanisms.
Suzuki et al.~\cite{suzuki2026} identified that failures typically occur during the final stages of placement, suggesting a need for OOD detection during high-precision manoeuvres.
Unlike these prior works that rely on a single policy, our approach introduces an ensemble-based policy that monitors action uncertainty across task manifolds to recognise and mitigate potential failures.

\subsection{Uncertainty Estimation and Safety Intervention in Robotic Manipulation}

Reliable runtime failure detection is a prerequisite for deploying IL policies in safety-critical chemistry laboratories. 
Recent research has focused on quantifying uncertainty to detect OOD states without requiring explicit failure data.
Romer et al. proposed FIPER~\cite{romer2025fiper}, which uses random network distillation to identify OOD observations. 
However, such visual-heavy embeddings can be confounded by the optical challenges of real-world chemistry labs, such as transparent glassware and solvents.
Similarly, FAIL-Detect~\cite{xu2025F} employs conformal prediction to calibrate safety bands, but its reliance on temporal consistency makes it sensitive to temporal drift caused by adaptive velocity scaling or a robot approaching its physical limits. 
Uncertainty quantification has also been explored using ensemble methods. 
Zhu et al.~\cite{Zhu2024} proposed a heterogeneity uncertainty quantification method for autonomous navigation through crowded intersections, distinguishing between data (aleatoric) and model (epistemic) uncertainties. 
While effective for 2D lateral and longitudinal control, these methods are primarily designed for low-dimensional action spaces.
Our work distinguishes itself from these approaches by treating failure prevention as a pre-emptive uncertainty-quantification problem. 
While existing systems like FAIL-Detect are often reactive, \textit{i.e.}, they identify a failure only after the state has significantly deviated, our \textbf{SAFE-CHEM} framework monitors epistemic drift to trigger a safety intervention before an irreversible event occurs.
Crucially, our method does not require any failure modes to be shown to the robot during training as the failure detection is handled via ensemble variance, which can be continually refined offline as the policy execution improves.

\begin{figure*}[!t]
    \centering
    \includegraphics[width=0.9\textwidth]{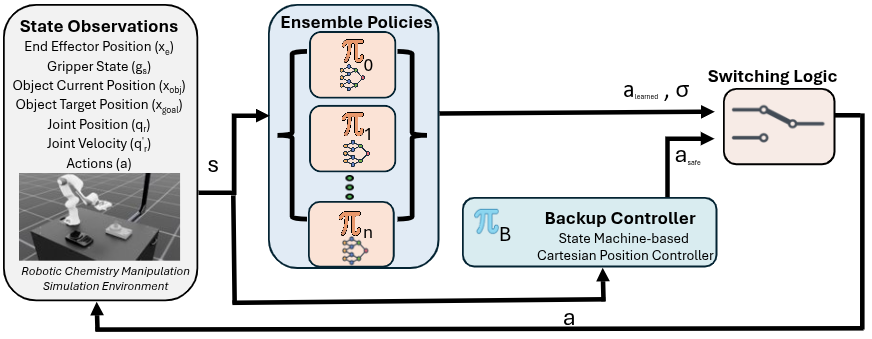}
    \caption{Overall system block diagram. Our framework demonstrates how a learned ensemble of RNN-based policies provide real-time action predictions ($a_{learned}$) and epistemic uncertainty quantification ($\sigma$). A deterministic state-machine-based backup controller ($\pi_B$) uses an explainable state machine for safe task completion. The switching logic autonomously selects the appropriate control signal ($a_{safe}$) based on a calibrated uncertainty threshold.}
    \label{fig:overall_block_diagram}
\end{figure*}
\section{Methodology}
% \subsection{Generating Training Simulation Environments +Data}
% \input{sections/methodology/environment_generation}
% \input{sections/methodology/creating_demos}
% \color{red}

% Here we need to introduce the task and motivation for choosing the architecture for the next subsections]
%% Intro to the methods sections
\par This work introduces \textbf{SAFE-CHEM}, an uncertainty-aware robotic skill framework based on deep imitation learning for chemistry laboratory automation (Section~\ref{ssec:ILmethod}).
As illustrated in Fig.~\ref{fig:intro_image} and Fig.~\ref{fig:overall_block_diagram}, our method uses a probabilistic ensemble of deep neural networks to capture the predictive distribution across multiple models trained for the same task, subsequently aggregating their outputs into a single, robust action (Section~\ref{ssec:UQmethod}).
This ensemble-based approach enables the estimation of epistemic uncertainty to identify OOD scenarios where the learned policy exhibits low confidence. 
To maintain operational safety, a threshold-triggered switching mechanism transitions control from the learned policy to a deterministic, rule-based fail-safe system whenever uncertainty exceeds a predefined limit to safeguard task integrity (Section~\ref{ssec:switching_method}).

\subsection{Imitation Learning for Robotic Manipulation in Chemistry Lab Automation}
% The use of learning methods for manipulator control is increasingly popular as these are able to generalise to a range of tasks and do not require hand crafted control mechanisms. 
\label{ssec:ILmethod}
%%Imitation Learning Formulation
Imitation learning methods have proven to be effective for training robots to perform complex tasks from expert human demonstrations~\cite{Pomerleau1989}.
We formalise the robot manipulation task as a discrete-time Markov decision processs (MDP), defined by the tuple $\mathcal{M} = (\mathcal{S}, \mathcal{A}, \mathcal{P}, R, \gamma, \rho_0)$, where $\mathcal{S}$ represents the continuous state space, $\mathcal{A}$ represents the continuous action space, $\mathcal{P}(s_{t+1} | s_t, a_t)$ denotes the state transition distribution, $R(s, a, s')$ is the reward function, $\gamma \in [0, 1)$ is the discount factor and $\rho_0$ is the initial state distribution.
%%Behaviour Cloning
\par In complex real-world chemistry laboratory environments, the reward function is difficult to analytically specify or sparse in nature.
Consequently, we utilise behaviour cloning~\cite{Pomerleau1989} to learn a policy $\pi_{\theta}$ directly from a dataset $\mathcal{D} = \{\tau_i\}_{i=1}^{N}$ consisting of $N$ expert demonstrations.
Each demonstration $\tau_i = \{(s_0, a_0), (s_1, a_1), \dots, (s_T, a_T)\}$ consists of state-action pairs that represent successful task completion.

For continuous action spaces, the objective is to find the optimal policy parameters $\theta^*$ by minimising the expected discrepancy between the expert's actions $a$ and the policy's predicted actions $\pi_{\theta}(s)$. 
This is formulated as the minimisation of the expected mean squared error, given by Equation~\ref{eq:mse_bc}, where where $\mathbb{E}_{(s, a) \sim \mathcal{D}}$ denotes the expectation over the distribution of state-action pairs $(s, a)$ contained within the expert dataset $\mathcal{D}$, and $\| \cdot \|_2^2$ represents the squared $L_2$ norm.

\begin{equation}
    % \theta^* = \arg\min_{\theta} \mathbb{E}{(s, a) \sim \mathcal{D}} \left[ | \pi{\theta}(s) - a |^2 \right]
    \theta^* = \arg \min_{\theta} \mathbb{E}_{(s, a) \sim \mathcal{D}} \left[ \| \pi_{\theta}(s) - a \|_2^2 \right]
    \label{eq:mse_bc}
\end{equation}

%% RNN
\par Vanilla BC assumes the Markov property, where the action $a_t$ depends solely on the current state $s_t$.
However, most robot manipulation tasks are inherently partially observable and require historical context to resolve ambiguities. 
To model these temporal correlations, we employ a recurrent neural network architecture for our policy.
Under this framework, the BC-RNN maintains an internal hidden state $h_t$ that serves as a compressed, latent representation of the interaction history.
This latent state is updated recursively at each timestep $t$ using a recurrent cell, such that $h_t = f(h_{t-1}, s_t)$.
The state $h_t$ is updated at each timestep $t$ using a recurrent cell.
The policy $\pi_\theta$ then maps this context-aware representation to the action space, $a_t = \pi_\theta(h_t)$.

% The goal of the behavior cloning trained agent  is to maximise the reward over the trajectory $T$ as shown in eq.\ref{eq:rewardTraj}
% \begin{equation}
%     \sum^{\inf}_{t=0} \gamma_tR(s_t, a_t, s_{t+1})\label{eq:rewardTraj}
% \end{equation}

% For behaviour cloning,  is collected where each demonstration is a trajectory that begins in the initial state $s_0^i \sim \rho_0$ and finishes in goal state $s_{T_i}^i \sim \rho_0$ defined in eq.\ref{eq:im_traj_def}.
% \begin{equation}
%     \tau_i = (s_0^i, a_0^i,s_1^i, a_1^i...s_{T_i}^i) \label{eq:im_traj_def}
% \end{equation}
 
\subsection{Uncertainty Quantification via Ensemble Variance}
% \par [Here we need to also talk about the ensemble network]

% \par ... using an ensemble of neural networks to generate a distribution of actions with the mean action and epistemic uncertainty
\label{ssec:UQmethod}
\par A standard policy is insufficient for safety-critical chemistry laboratory environments, where model overconfidence in novel or high-risk states could be catastrophic.
To mitigate these risks, we employ an ensemble of $M$ neural networks $\{\pi_\theta\}_{m=1}^M$, which quantifies the epistemic uncertainty of the model.
Each ensemble member is independently initialised and trained on the expert dataset $\mathcal{D}$, effectively approximating a Bayesian posterior over the policy parameters.
For a given latent temporal state $h_t$, each member $m$ generates an action prediction $a_{t, m} = \pi_{\theta_m} (h_{t, m})$.
The final control signal $\bar a_t$ is derived from the predictive mean, where $\bar{a}_t = \frac{1}{M} \sum\limits_{m=1}^{M} a_{t,m}$.
Our ensemble also captures the temporal divergence of the recurrent latent states through the variance $\sigma_e^2$.
We hypothesise that the the ensemble's disagreement correlates with the expected prediction error; specifically, as the robot encounters states further from the expert's training manifold, the inter-model variance should increase proportionally with the MSE.

% The uncertainty here is represented by the variance $\sigma$.

\subsection{Uncertainty-Aware Switching Mechanism}
\label{sec:uncertainty_switching_method}
\label{ssec:switching_method}
Building upon the uncertainty quantification derived from ensemble variance, we propose a switching mechanism that translates probabilistic signals into discrete control transitions. 
The design is predicated on the stochastic divergence of the action ensemble between nominal and failure (unsafe) states. 
By modelling the density of action variances observed during successful task executions, we define the boundaries of the agent’s competence manifold.
To distinguish between nominal variance and variance indicative of impending failure we calibrate component-specific thresholds using a set of reference trajectories.
During task execution, we consider the trajectory data for each action component $\mathbf{a}_i \in \mathbb{R}^n$, where $n = 7$ which denotes the degrees-of-freedom (DoF) of the robotic manipulator.
For each rollout $\tau_k$, we monitor the ensemble variance $\sigma^2_{i,t}$ at each timestep $t$ for component $i$ using Equation~\ref{eq:variance}.
Here, $\mu_{i,t}$ represents the mean action value.

\begin{equation}
\sigma^2_{i,t} = \mathbb{E}[(a_{i,t} - \mu_{i,t})^2]
\label{eq:variance}
\end{equation}

To standardise trajectories of variable durations, zero-padding is applied to reach a maximum horizon $H$.
We classify each trajectory $\tau_k$ based on task outcome as either complete (successful) ($\tau_k \in \mathcal{C}$) or failed ($\tau_k \in \mathcal{F}$).
% For each successful trajectory, we analyse the variance $(\sigma^2)$ distributions for each action component to establish component-specific confidence thresholds. 
We aggregate all non-zero variance values from trajectories in each class to define the reference sets.

\begin{equation}
V^{(\mathcal{C})}_i = \{\sigma^2_{i,t} \mid \sigma^2_{i,t} > 0, \, \tau \in \mathcal{C}, \, t \in [1, H]\}
\end{equation}

\begin{equation}
V^{(\mathcal{F})}_i = \{\sigma^2_{i,t} \mid \sigma^2_{i,t} > 0, \, \tau \in \mathcal{F}, \, t \in [1, H]\}
\end{equation}

% The design of the switching mechanism is predicated on the stochastic divergence of the action ensemble between nominal and failure states.
% This behaviour is characterised by modelling the density of action variances observed during successful task executions, which defines the boundaries of the agent's competence manifold.
We model the underlying distribution of successful variances using kernel density estimation (KDE) to provide a smooth, non-parametric estimate of the success-conditioned density.
% , we characterise the underlying distribution of these variances to derive a continuous, non-parametric estimate of the success-conditioned density.
% we derived a distinct shift toward lower variances for successful trajectories compared to failed ones. 

% To distinguish between nominal variance and variance indicative of impending failure, we calibrate our thresholds using a set of reference trajectories. 
% We define $V_i^{(\mathcal{S})}$ as the set of all non-zero variance values observed during successful task executions $\mathcal{S}$:

% \begin{equation} 
% V^{(\mathcal{S})}_i = \{ \sigma^2_{i,t} \mid \sigma^2_{i,t} > 0, \tau \in \mathcal{S}, t \in [1, H] \} 
% \end{equation}

% We model the underlying distribution of successful variances using kernel density estimation (KDE) to provide a smooth, continuous estimate of the success density $\hat{f}_{\mathcal{S}}(v)$:

% \begin{equation} 
% \hat{f}_{\mathcal{S}}(v) = \frac{1}{|V^{(\mathcal{S})}_i|} \sum_{v_j \in V^{(\mathcal{S})}_i} K_h(v - v_j) \label{eq:kde} 
% \end{equation}
The density is estimated using a Gaussian kernel $K_h(u)$ with bandwidth $h$ via Scott's Rule \cite{silverman2018density}. 
\begin{equation}
    K_h(u) = \frac{1}{\sqrt{2\pi}h}\exp\left(-\frac{u^2}{2h^2}\right)
\end{equation}
By integrating this density, we obtain the cumulative distribution function (CDF), $F_{\mathcal{C}}(v) = P(\sigma^2 \leq v \mid \tau \in \mathcal{C})$.
% = \int_{0}^{v} \hat{f}_{\mathcal{S}}(u) \, du 
% \label{eq:cdf} 
% \end{equation}
For a specified quantile $q \in (0,1)$, we define the component-specific confidence threshold $\theta_{i,q}$ as the value below which $q \times 100\%$ of successful variances reside:

\begin{equation} 
\theta_{i,q} = \inf \{ v : F_{\mathcal{C}}(v) \geq q \} \label{eq:threshold} 
\end{equation}

% For the switching logic, we implement a sliding window peak detection method. 
%that identifies sustained periods of high variance. 
During online execution, the system monitors the ensemble variance $\sigma^2_{i, t} = \mathbb{E}[(a_{i,t} - \mu_{i,t})^2]$ for each action component. 
To prevent false-positive transitions triggered by transient noise or momentary spikes in uncertainty, we implement a sliding window peak detection algorithm. This mechanism ensures that a switch to the safety backup controller occurs only during sustained periods of high variance.

For a given trajectory $\tau$ and action component $i$, the detection signal $D_{i,t}(\tau)$ is defined as positive at timestep $t$ if the number of instances where the variance exceeds the calibrated threshold $\theta_{i,q}$ within a window of size $w$ reaches a limit $n_{\text{peaks}}$:

% \begin{equation}
% D_{i, t}(\tau) = \mathbb{I} \left[\sum_{j=t-w}^{t} \mathbb{I}(\sigma^2_{i,j} \geq \theta_{i,q}) \geq n_{\text{peaks}}\right]
% \label{eq:detection}
% \end{equation}

\begin{equation}
D_{i, t}(\tau) = \left[\sum_{j=t-w}^{t} [\sigma^2_{i,j} \geq \theta_{i, q}] \geq n_{\text{peaks}}\right]
\label{eq:detection}
\end{equation}

% where $\mathbb{I}[\cdot]$ denotes the indicator function. 
If $D_{i,t}(\tau) = 1$ for any action component $i \in \{1, \dots, n\}$, the switching logic flags an uncertainty risk. 
If more than two action components are flagged, the system transitions to the backup policy ($\pi_B$) shown in Fig.~\ref{fig:overall_block_diagram}.
The sensitivity of the detection performance depends on the hyperparameters $w$ and $n_{\text{peaks}}$, which are empirically tuned to balance safety with task completion. 

% The complete procedure for the runtime switching logic is detailed in Algorithm~\ref{alg:switching_logic}.

% \input{sections/methodology/confidence_interval}
% \input{sections/methodology/switching_logic}

\section{Experimental Evaluation}
\label{sec:experimental_eval}

We evaluated our uncertainty-aware imitation learning framework across three chemistry-based manipulation tasks (Fig.~\ref{fig:task_setups}) to address the following: (1) Does transitioning from a single policy to a distributed ensemble improve task success? (2) Can KDE effectively model the stochastic variance of an ensemble to define a quantile-based triggering mechanism for intervention? and (3) How does the framework balance the efficiency of learning-based policies with the reliability of rule-based safety controllers to maximise overall success?
Finally, we demonstrate the zero-shot transfer capabilities of our method on a real robotic system.

\subsection{Experimental Setup}
\label{ssec:exp_setup}
All simulation environments were developed within IsaacLab~\cite{mittal2025, Darvish2025}.
These were run on a system featuring an AMD Ryzen Threadripper 7970X processor, 128GB of RAM and an NVIDIA GeForce RTX 5090 GPU, operating on Ubuntu 24.04. 
For consistency across domains, a Franka Production 3 (FP3) robotic manipulator~\cite{Haddadin2022} was used in both simulated and real-world configurations.
Our learning framework employed an ensemble of BC-RNN models from the RoboMimic framework~\cite{robomimic2021} to effectively capture the temporal dynamics of the manipulation task. 
We use BC as our policy backbone because its low computational overhead and high inference frequency are better suited for the online requirements of robotic chemists than more resource-intensive architectures, such as diffusion policies.
The network observes a low-dimensional state vector $\mathcal{O}\in \mathcal{R}^{29}$ encompassing the robot's proprioceptive data (end-effector position ($x_e \in \mathcal{R}^7$), gripper state ($g_s \in \mathcal{R}^2$), joint position ($q_r \in \mathcal{R}^7$) and velocity ($\dot q_r \in \mathcal{R}^7$) states,) and environmental state (object current ($x_{obj}\in \mathcal{R}^3$ ) and target ($x_{goal}\in \mathcal{R}^3$)) states. We chose these observations to obtain a balance between efficient learning and reducing computational cost. 
Temporal dependencies are processed using a two-layer long short-term memory (LSTM) architecture with a hidden dimension of 400, trained with a sequence length of 16. 
The data generation pipeline was run for each task separately and begins with 10 human demonstrations collected in simulation using a SpaceMouse recording the state and action pairs for each trajectory. 
We used MimicGen~\cite{mandlekar2023mimicgen} to annotate the subtasks (such as grasping the object) to prepare the dataset for demonstration generation. 
These annotated demonstrations were expanded using the data generation tools in MimicGen to produce a dataset of 4,000 valid trajectories per task. 
During training to account for human demonstration multi-modality, the continuous action space is modelled using a Gaussian mixture model (GMM) with 10 modes. 
To estimate epistemic uncertainty and improve robustness, we utilise a deep ensemble approach comprising 15 distinct BC-RNN models. 
To maximise ensemble diversity, each model is initialised with a unique random seed and trained on a randomly sampled 25\% subset of the total demonstration dataset so each policy is trained with a dataset of 1000 demonstrations. 
The models are trained using the Adam optimiser with an initial learning rate of $1e^{-4}$ and a batch size of 100 for 1000 epochs.

\subsection{Laboratory Tasks}
\label{ssec:lab_tasks}

\begin{figure}[t]
    \centering
    \begin{subfigure}[t]{0.15\textwidth}
        \centering
        \includegraphics[width=\textwidth]{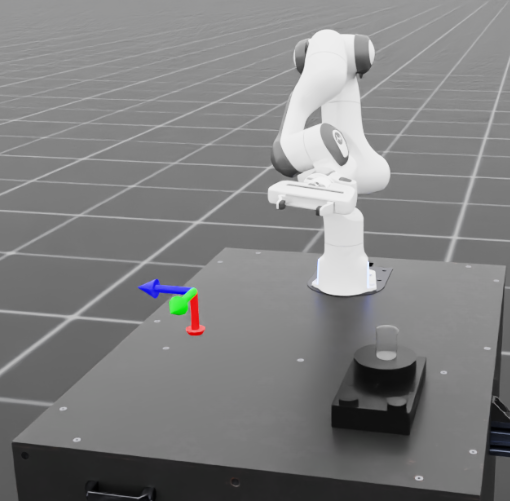}
        \caption{\textbf{Lift}}
        \label{fig:image_a}
    \end{subfigure}
    \hfill
    \begin{subfigure}[t]{0.15\textwidth}
        \centering
        \includegraphics[width=\textwidth]{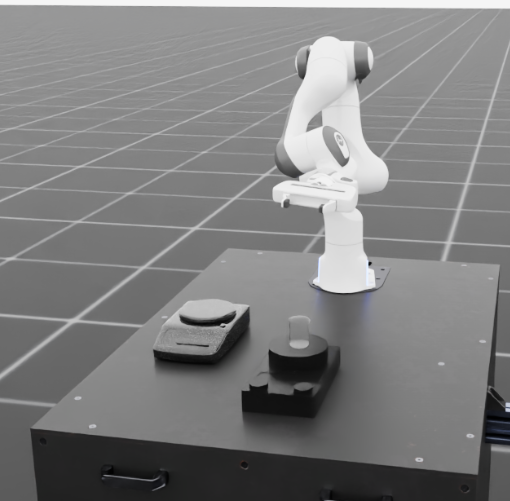}
        \caption{\textbf{Pick and Place}}
        \label{fig:image_b}
    \end{subfigure}
    \hfill
    \begin{subfigure}[t]{0.15\textwidth}
        \centering
        \includegraphics[width=\textwidth]{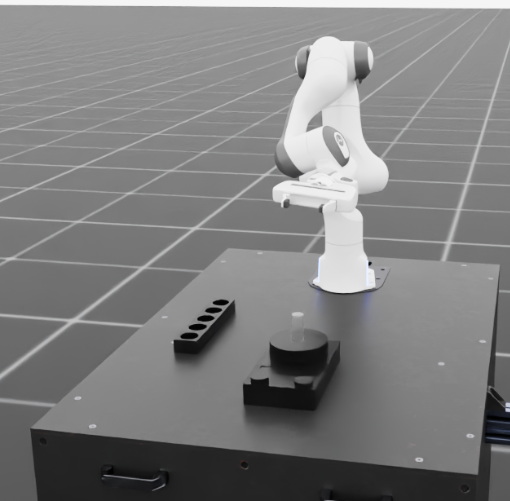}
        \caption{\textbf{Insertion}}
        \label{fig:image_c}
    \end{subfigure}
    % \hfill
    % \begin{subfigure}[t]{0.19\textwidth}
    %     \centering
    %     \includegraphics[width=\textwidth]{example-image-a}
    %     \caption{Real robot setup}
    %     \label{fig:image_d}
    % \end{subfigure}
    % \hfill
    % \begin{subfigure}[t]{0.19\textwidth}
    %     \centering
    %     \includegraphics[width=\textwidth]{example-image-a}
    %     \caption{Heating and Weighing}
    %     \label{fig:image_e}
    % \end{subfigure}
    \caption{The robotic chemistry manipulation simulated tasks.}
    \label{fig:task_setups}
\end{figure}

The experimental evaluation is conducted across three manipulation tasks that represent core operations in an automated chemistry laboratory determined from studying the types of activities that chemists performed in research laboratory (Fig.~\ref{fig:task_setups}).
In the \textbf{lift task}, the robot picks up an object (\textit{e.g.}, a vial or a beaker) and hold it in a desired position. 
This task is designed for visual inspection tasks in chemistry experimental workflows~\cite{Walker2023} where chemists may inspect vial samples for colour change or precipitation. 
In the \textbf{pick and place task}, the robot picks up an object (\textit{e.g.,} a vial or a beaker) from an instrument (\textit{e.g.,} a scale) and places it on another instrument (\textit{e.g.,} IKA RCT plate). 
The pick and place task is the most predominant manipulation skill in automated chemistry workflows~\cite{burger2020, lunt2024}, for example after weighing out a powder and solute and placing on an IKA RCT plate to heat and stir to dissolve the solution.
In the \textbf{insertion task}, the robot picks up a vial from an instrument (\textit{e.g.,} IKA RCT plate) and places it inside a rack for transportation or further analysis. 
This task is a high precision, contact-rich manipulation skill that is typically found prior to sample transportation in end-to-end experimental workflows~\cite{Longley2026}.
In our lift and pick and place tasks, the robot needs to adhere to chemists' preferences of grasping the objects from the side rather than from the top as typically seen in object manipulation tasks. 
This adds complexity to the task as the robot is consistently working closer to the joint limits in this grasp pose as opposed to the top down grasp pose. 
We also introduce domain randomisation in the environment in order to allow the trained policies to generalise to a range of initial and goal states. 
The object start position and goal position were randomly assigned within a $0.2m \times 0.2m$ region in the environment during all stages of the workflow.  

% The experiments include an evaluation of the ensemble of policies, the impact of the confidence level for intervention, evaluation on unknown objects and finally a demonstration of the control approach on the real robot. 
\subsection{Experiment I: Ensemble Policy Evaluation}
\label{ssec:ensemble_policy_eval}
To evaluate the ensemble network's performance, we measure the task success rate and the safety incident rate, with the latter defined as any instance where an object is dropped or tilted beyond $45^0$. 
For our experiment we evaluate the ensemble of policies running in an environment with a step size of $0.01s$ ($100Hz$).
With an action repeat of two, the physics engine executes two simulation steps per control cycle, yielding an effective policy control frequency of $50Hz$.
We conducted $3 \times 100$ evaluation episodes per task, reporting the mean and standard deviation to account for stochasticity in object spawn locations.
% This reduces the influence of the random nature of the object spawn location.
% The task is classified as failed if the object is dropped or tilted beyond $\angle{45}$ as this would classify as a safety incident. 
We first investigate the impact of using an ensemble of policies on the success rate of the tasks. 
We evaluate the rollout of the task with an ensemble size of 1 (single policy), 5, 10 and 15 over a maximum horizon of 5,000 steps. 
Table~\ref{tab:ensemble_size} summarises the relationship between ensemble size and task success rate.
It demonstrates that employing an ensemble of policies significantly enhances success rates across all evaluated tasks.
We observe distinct uncertainty profiles: for lift and pick and place tasks, epistemic uncertainty peaks during the initial grasping phase, whereas the insertion task exhibits maximum uncertainty toward the end of execution due to the high precision required for vial alignment.
While performance gains did not scale linearly with ensemble size, an optimal size of at least $n = 10$ was consistent across tasks.
For the lift task, diminishing returns were observed beyond $n=10$. 
Conversely, the pick and place task required $n=15$ to achieve a significant performance leap, likely because a larger ensemble is needed to effectively mitigate the influence of outlier predictions during the placing phase. 
For the insertion task, the policy showed no improvement beyond $n = 10$, and performance actually declined significantly for the ensemble size of $n=15$. 
This suggests that larger ensembles may produce overly conservative action predictions, leading to smaller control outputs and subsequent task timeouts.

% \begin{table}[h]
% \centering
% \begin{tabular}{c|c|c|c}
% \textbf{Size} & Lift ($\% \pm SD$)  & Place ($\% \pm SD$) & Insert ($\% \pm SD$) \\ \hline
% 1                      & 75.7$\pm$2.5  & 67.7$\pm$9.1   & 21.3$\pm$0.6\\ \hline
% 5                      & 88.3$\pm$4.2  & 63.7$\pm$7.6   & 51.7$\pm$4.0  \\ \hline
% 10                     & \textbf{93.0$\pm$1.7}  & 69.0$\pm$8.5   & \textbf{55.0$\pm$9.8} \\ \hline
% 15                     & 92.0$\pm$1.0  & \textbf{83.0$\pm$2.7}   & 42.0$\pm$4.0
% \end{tabular}
% \caption{Success rate and standard deviation for the of $3\times 100$ rollouts of the task policies for a range of ensemble sizes for the lift, place and insertion tasks}
% \label{tab:ensemble_size}
% \end{table}

\begin{table}[h]
\centering
\caption{Average task success rate (\%) and standard deviation (SD) for varying ensemble sizes across all simulated tasks.}
\label{tab:ensemble_size}
\small
\resizebox{\columnwidth}{!}{%
\begin{tabular}{@{} c *{3}{c} @{}}
\toprule
\textbf{Size} & \textbf{Lift} ($\% \pm \text{SD}$) & \textbf{Pick and Place} ($\% \pm \text{SD}$) & \textbf{Insertion} ($\% \pm \text{SD}$) \\ 
\midrule
1  & $75.7\pm2.5$ & $67.7\pm9.1$ & $21.3\pm0.6$ \\
5  & $83.3\pm4.2$ & $63.7\pm7.6$ & $51.7\pm4.0$ \\
10 & $\mathbf{93.0\pm1.7}$ & $69.0\pm8.5$ & $\mathbf{55.0\pm9.8}$ \\
15 & $92.0\pm1.0$ & $\mathbf{83.0\pm2.7}$ & $42.0\pm4.0$ \\ 
\bottomrule
\end{tabular}%
}
\end{table}

% During the lift and place tasks, we observe that the period of the task with the highest uncertainty is the grasping step, however in the insert task the uncertainty is greatest towards the end of the task execution due to the high level of manipulation precision required. 
% We did not observe a consistent improvement in task completion with increasing ensemble size, however the optimal ensemble size for all the tasks investigated were at least 10, suggesting that for most tasks and ensemble size of 10 or greater will lead to optimal task completion.
% In the simplest task (lift), the increasing the number of policies collaborating does not provide significant benefit above an ensemble size of 10, whereas for the pick and place task, the impact of the increased ensemble size is not significant until we reach 15 policies. 
% This may be due to an outlying place policy that is not effectively buffered until the ensemble size reaches 15. 
% For the insertion task, the addition of policies above an ensemble size of 5 did not show an improvement in the task success rate and above an ensemble of 10, the task performance decreased. 
% This may be due to the larger ensemble of polices predicting a more conservative action to be taken, tending to cause smaller action to be taken and the task to timeout. 

\subsection{Experiment II: Calibrating the Switching Logic}
\label{ssec:calibrate_switch_logic}
\begin{figure}[h]
    \centering
    \includegraphics[width=0.99\linewidth]{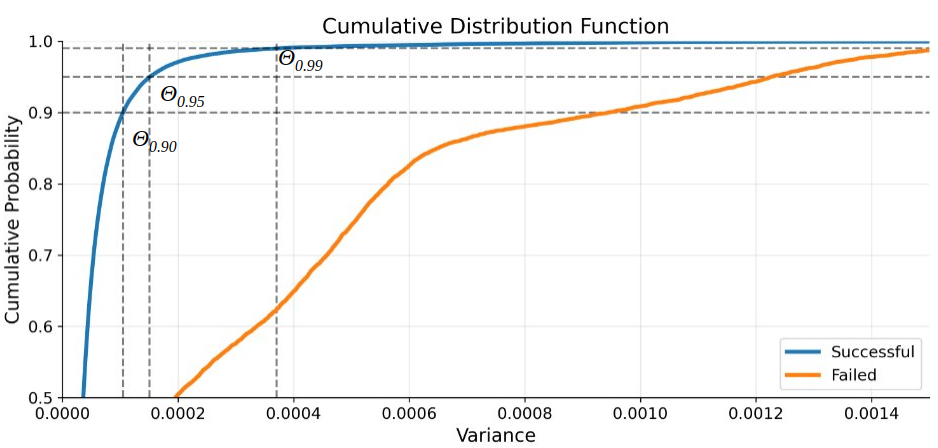}
    \caption{Confidence thresholds calculated at the $\theta_{0.90}$, $\theta_{0.95}$ and $\theta_{0.99}$ bounds.}
    \label{fig:ci_intervals}
\end{figure}

% To improve the implementation of learned policies, we propose a simple, lightweight data-driven approach to identify critical intervention points during robotic manipulation tasks by analysing action space variance distributions across successful and failed trajectory executions. 
We evaluate the proposed switching logic on a 7-DoF robotic manipulator across three different manipulation tasks.
To calibrate the mechanism, we initially collected trajectory data from ensemble policy rollouts without intervention, classifying each as completed (successful) or failed. 
% During task execution, we collect trajectory data for each action component $\mathbf{a}_i \in \mathbb{R}^7$ representing the 7-DOF robotic manipulator. 
As established in Section~\ref{sec:uncertainty_switching_method}, we applied KDE to the aggregated variance data.
Empirical observations (Fig.~\ref{fig:ci_intervals}) show a distinct shift toward lower variances for successful trajectories ($\mathcal{C}$) compared to failed ones ($\mathcal{F}$), validating the assumption that the ensemble policy maintains higher consistency during confident, successful actions.
We investigated three confidence thresholds at specified quantiles $\theta_q$ for a conservative $(q = 0.90)$, moderate $(q = 0.95)$ and relaxed $(q = 0.99)$ uncertainty limits.

Due to certain actions such as grasping being inherently uncertain, we trigger a switch only when the condition in Equation~\ref{eq:detection} is met by more than two action components. 
We utilised a sliding window size $w\in [8,10]$ and conducted a grid search over $n_{\text{peaks}} \in \{1, \dots, 10\}$.
Detection performance was measured via detection rate (true positives), false alarm rate (false positives), and F1-Score. 
As shown in Table~\ref{tab:cal_switching_logic}, an $n_{\text{peaks}}$ value of 8 consistently provided the highest F1-score across lift, pick and place, and insertion tasks when using the conservative $\theta_{0.90}$ limit. 
This configuration was selected for final implementation to maximise failure detection while minimising unnecessary interruptions. A longer window with results in a delayed detection of policy uncertainty and increased occurrence of false positive uncertainty detection whereas a shorter window tends to reduce the false positive uncertainty detections but is more likely to miss short but critical periods of higher uncertainty.

\begin{table}[h]
\centering
\caption{Peak detection algorithm  performance across a range of window sizes $w$ and peak counts $n_{\text{peaks}}$.}
\label{tab:cal_switching_logic}
\resizebox{\columnwidth}{!}{%
\begin{tabular}{@{} c *{3}{c} @{}}
\toprule
$\mathbf{n_{\textbf{peaks}}}$ & \textbf{Lift} ($w=10$) & \textbf{Place} ($w=10$) & \textbf{Insert} ($w=9$) \\ 
\midrule
1  & 0.67 & 0.67 & 0.67 \\
2  & 0.67 & 0.67 & 0.67 \\
3  & 0.67 & 0.67 & 0.67 \\
4  & 0.71 & 0.67 & 0.69 \\
5  & 0.76 & 0.70 & 0.74 \\
6  & 0.80 & 0.78 & 0.78 \\
7  & 0.84 & 0.86 & 0.82 \\
8  & \textbf{0.90} & \textbf{0.91} & \textbf{0.87} \\
9  & 0.88 & 0.90 & \textbf{0.87} \\
10 & 0.84 & 0.89 & 0.85 \\ 
\bottomrule
\end{tabular}%
}
\end{table}

\subsection{Experiment III: Confidence Level Intervention}
\label{ssec:CI_intervention}
We evaluate task success rates at different intervention levels ($\theta_q$) within the simulation environment.
First we find the most conservative confidence level of $\theta_{0.90}$ of the successful task uncertainties and the window that best distinguishes between the successful and failed task executions for each task. 
As part of the hybrid architecture, we implemented a parallelised, finite-state machine (FSM) backup controller. 
The controller is implemented utilising NVIDIA Warp~\cite{NvidiaWarp2024} to allow parallelisation across $N$ simulation environments directly on the GPU. 
We define the distinct operational phases of each of the task and the start and end conditions of these phases.
Although these rule-based controllers exhibit longer task completion times than the learned policies, they ensure reliable task progress independent of task uncertainties. 
The overall performance of this hybrid uncertainty-aware framework is summarised in Table~\ref{tab:task_success_rate} 
%the intervention rate of the switching logic is shown in Table~\ref{tab:intervention_rate} 
and the critical failure rate is shown in Table~\ref{tab:safety_failure_rate}. 
%Task Success Rate Table 
% \begin{table*}[t]
% \begin{tabular}{l|llll|llll|llll}
%                        & \multicolumn{4}{c|}{Success rate (lift)} & \multicolumn{4}{c|}{Success rate (place)} & \multicolumn{4}{c}{Success rate (insert)} \\
% \textbf{Size} & None & $\theta_{0.90}$ & $\theta_{0.95}$ & $\theta_{0.99}$ & None & $\theta_{0.90}$ & $\theta_{0.95}$ & $\theta_{0.99}$ & None & $\theta_{0.90}$ & $\theta_{0.95}$ & $\theta_{0.99}$  \\ \hline
% 1                      & 75.7$\pm2.5$    & -        & -        & -       & 67.7$\pm9.0$   &          & -        & -        &  21.3$\pm$0.6  &     -    &     -     &      -    \\
% 5                      & 83.3$\pm$4.2 & 96.0$\pm$2.0 & 97.7$\pm$2.1  & 96.3$\pm$ 1.5& 63.7$\pm$7.6 & 94.3$\pm$1.5 & \textbf{98.3$\pm$1.2}& 89.3$\pm$1.5 & 51.7$\pm$4.0  & 60.0$\pm$4.4 & 58.3$\pm$3.8 & 59.0$\pm$2.7 \\
% 10                     & 93.0$\pm$1.7 & 96.3$\pm$0.6 & 99.0$\pm$1.0 & \textbf{99.3$\pm$0.6 }  & 69.0$\pm$8.5 & 96.3$\pm$0.6 & 97.0$\pm$2.0 & 86.4$\pm$4.0   &   55.0$\pm$9.8        &   67.3$\pm$3.5      &      62.7$\pm$5.5    &     58.0$\pm$6.1     \\
% 15                     &  92.0$\pm$1.0& 98.0$\pm$1.0& 99.0$\pm$0.0  & 98.7$\pm$1.5   & 83.0$\pm$2.7 & 96.7$\pm$1.5 & 96.0$\pm$1.7  & 91.0$\pm$1.7    & 42.0$\pm$4.0     & \textbf{68.0$\pm$1.7}        &   64.3$\pm$4.2       & 59.7$\pm$2.5         
% \end{tabular}
% \caption{}
% \label{tab:task_success_rate}
% \end{table*}

\begin{table*}[t]
\centering
\caption{Task success rate (\%) across varying ensemble sizes and confidence thresholds $\theta_q$. \textit{None} indicates the baseline performance without the switching mechanism.}
\label{tab:task_success_rate}
\resizebox{\textwidth}{!}{%
\begin{tabular}{@{} l *{4}{c} | *{4}{c} | *{4}{c} @{}}
\toprule
 & \multicolumn{4}{c}{\textbf{Success rate (Lift)}} & \multicolumn{4}{c}{\textbf{Success rate (Pick and place)}} & \multicolumn{4}{c}{\textbf{Success rate (Insertion)}} \\
\cmidrule(lr){2-5} \cmidrule(lr){6-9} \cmidrule(l){10-13}
\textbf{Size} & None & $\theta_{0.90}$ & $\theta_{0.95}$ & $\theta_{0.99}$ & None & $\theta_{0.90}$ & $\theta_{0.95}$ & $\theta_{0.99}$ & None & $\theta_{0.90}$ & $\theta_{0.95}$ & $\theta_{0.99}$ \\
\midrule
1  & $75.7\pm2.5$ & -- & -- & -- & $67.7\pm9.0$ & -- & -- & -- & $21.3\pm0.6$ & -- & -- & -- \\
5  & $83.3\pm4.2$ & $96.0\pm2.0$ & $97.7\pm2.1$ & $96.3\pm1.5$ & $63.7\pm7.6$ & $94.3\pm1.5$ & $\mathbf{98.3\pm1.2}$ & $89.3\pm1.5$ & $51.7\pm4.0$ & $60.0\pm4.4$ & $58.3\pm3.8$ & $59.0\pm2.7$ \\
10 & $93.0\pm1.7$ & $96.3\pm0.6$ & $99.0\pm1.0$ & $\mathbf{99.3\pm0.6}$ & $69.0\pm8.5$ & $96.3\pm0.6$ & $97.0\pm2.0$ & $86.4\pm4.0$ & $55.0\pm9.8$ & $67.3\pm3.5$ & $62.7\pm5.5$ & $58.0\pm6.1$ \\
15 & $92.0\pm1.0$ & $98.0\pm1.0$ & $99.0\pm0.0$ & $98.7\pm1.5$ & $83.0\pm2.7$ & $96.7\pm1.5$ & $96.0\pm1.7$ & $91.0\pm1.7$ & $42.0\pm4.0$ & $\mathbf{68.0\pm1.7}$ & $64.3\pm4.2$ & $59.7\pm2.5$ \\
\bottomrule
\end{tabular}%
}
\end{table*}

\subsubsection{On Task Success Rate}
% task success rate 
Across all experiments, the integration of the uncertainty-aware switching logic consistently improved success rates, with the hybrid system outperforming the non-intervened ensemble in every instance.
For the lift task, the best performing model used an ensemble of 10 policies with the most relaxed confidence level ($\theta_{0.99}$), achieving an average success rate of $99.3\%$. 
Because larger ensembles ($n \geq 10$) exhibit high baseline performance on this simple task, the marginal improvement offered by the hybrid controller was less pronounced compared to its impact on smaller ensembles or single-policy baselines. This task is relatively simple and so the policy is easily able to handle completing the task to a high standard. 
The pick and place task also shows a significant improvement in the task success when using the hybrid method.
The highest success rate was observed with an ensemble of $n = 5$ and the moderate confidence interval ($\theta_{0.95}$), though performance at $n = 10$ remained within one standard deviation. 
Given the increased complexity of this task, the success rate for the learned policy alone was markedly lower than in the lift task, highlighting the utility of the intervention mechanism.
In the most challenging task, \textit{i.e.}, the vial insertion, we observed that the best task success rates were achieved with the most conservative threshold ($\theta_{0.90}$) for all ensemble sizes. 
The high-precision requirements of this task often led the learned policy toward failure; however, the conservative threshold triggered the backup controller more frequently, effectively compensating for policy instability.
Notably, we observed instances where the learned policy resulted in unrecoverable states that failed to trigger the switching mechanism. 
For example, a vial slightly misplaced in the rack might remain improperly seated without exceeding the calibrated uncertainty threshold.
Our current framework does not explicitly address such failures; consequently, the backup controller was occasionally unable to reset the vial.
As this task requires a high level of precision, future work will explore integrating multimodal sensing, such as force-torque feedback; this would allow the system to augment the uncertainty profile with force signatures and trigger the backup controller more reliably before failure states become unrecoverable.
% We observed during this task that the learned policy would sometimes create unrecoverable states that were not critical safety incidents. 
% These occurred when the vial was slightly misplaced in the vial rack meaning that the vial was in the vial rack but not completely inserted. 
% The backup controller was not able to reach the vial to grasp it and place correctly in the vial rack, leading to the attempt being failed but no safety incident occurred. 

% \subsubsection{On Switching Intervention Rate}
% % intervention rate 
% Our results demonstrate that the intervention rate of the switching logic reduces with the increasing uncertainty limit across most tasks. 
% At the most conservative $\theta_{0.90}$ threshold, larger ensembles exhibit higher intervention rates; however, as this limit increases, the intervention frequency for larger ensembles drops below that of smaller ones.
% For the challenging vial insertion task, intervention rates remained relatively consistent between the $\theta_{0.90}$ and $\theta_{0.95}$ for ensembles with fewer than 15 polices. 
% This suggests that for high-precision tasks, larger ensembles provide a more robust characterisation of epistemic uncertainty, allowing the switching logic to better distinguish between nominal and failure-prone actions.

\subsubsection{On Safety Failure Rate}
% safety failure rate 
For the lift task, using an ensemble of policies as rather than a single policy reduced the safety violation rate however simply increasing the number of policies in the ensemble did not significantly benefit the rate of safety violations. 
For the lift task, the ensemble of 10 policies did not cause any safety violations, suggesting that the unsafe actions were effectively mitigated by the ensemble. 
Qualitative observations during task execution revealed that non-intervened ensemble policies primarily failed due to timeouts, often as a result of the robot failing to initiate a grasping action. 
Conversely, under the hybrid control framework, the switching mechanism occasionally triggered at suboptimal moments, causing the backup controller to tip the object and inadvertently induce a safety violation.
However, an exhaustive collection of safety failure modes was not generated, and so we have only observed safety failures that have been predefined. 
The inherent complexity of the pick and place task elevates the risk of operational failures; a higher density of environmental obstacles increases collision probabilities, while the addition of the placement phase introduces further susceptibility to object drops or instability.
We observed that with the added complexity in this pick and place task, we saw more impact on the reduction of the safety violation rate by using the ensemble of policies. 
For the insertion task where the potential for safety failures is particularly high, we observe a reduction in the safety failure rate by using the ensemble of policies. 
With the single policy alone, around 16\% of the failures were due to a timeout, with all remaining failed task executions resulting from the policy causing a safety failure. 
By using the ensemble of policies, the proportion of failures that were due to safety incidents was reduced to less than half of all failures being due to safety failures. 
The insertion task did not benefit from the hybrid approach in regards to the safety failure rate as in the other tasks. 
During execution, manipulation failures predominantly occurred when the robot attempted to insert the vial into the rack at an angular offset from the vertical axis. 
Because the rack's support structure encompasses only half the height of the vial, premature release typically caused the vial to topple out of the holder if the placement alignment was imperfect. 
Because these dynamics resulted in an irreversible environmental state, the task could not be completed, irrespective of backup controller intervention.

\begin{table*}[t]
\centering
\caption{Rate of critical safety violations (\%) (\textit{e.g.}, dropped objects or tilts $> 45^\circ$) across varying ensemble sizes and confidence thresholds $\theta_q$.}
\label{tab:safety_failure_rate}
\resizebox{\textwidth}{!}{%
\begin{tabular}{@{} l *{4}{c} | *{4}{c} | *{4}{c} @{}}
\toprule
 & \multicolumn{4}{c}{\textbf{Safety failure rate (Lift)}} & \multicolumn{4}{c}{\textbf{Safety failure rate (Pick and place)}} & \multicolumn{4}{c}{\textbf{Safety failure rate (Insertion)}} \\
\cmidrule(lr){2-5} \cmidrule(lr){6-9} \cmidrule(l){10-13}
\textbf{Size} & None & $\theta_{0.90}$ & $\theta_{0.95}$ & $\theta_{0.99}$ & None & $\theta_{0.90}$ & $\theta_{0.95}$ & $\theta_{0.99}$ & None & $\theta_{0.90}$ & $\theta_{0.95}$ & $\theta_{0.99}$ \\
\midrule
1  & $3.3\pm0.6$  & -- & -- & -- & $11.3\pm5.0$ & -- & -- & -- & $62.3\pm5.1$ & -- & -- & -- \\
5  & $1.7\pm2.9$  & $1.0\pm1.0$ & $0.3\pm0.6$ & $3.7\pm1.5$ & $8.0\pm1.0$  & $4.0\pm2.0$ & $1.7\pm1.2$ & $2.0\pm1.7$ & $25.3\pm1.2$ & $33.0\pm3.6$ & $34.7\pm1.5$ & $33.3\pm1.2$ \\
10 & $0.0\pm0.0$  & $0.0\pm0.0$ & $0.7\pm0.6$ & $0.7\pm0.6$ & $6.0\pm3.0$  & $1.7\pm0.6$ & $3.0\pm2.0$ & $3.0\pm1.0$ & $21.0\pm7.0$ & $24.7\pm2.3$ & $26.7\pm4.0$ & $29.3\pm3.1$ \\
15 & $2.3\pm1.5$  & $1.3\pm0.6$ & $1.0\pm0.0$ & $1.3\pm1.5$ & $3.0\pm1.0$  & $2.0\pm1.0$ & $2.7\pm1.2$ & $2.0\pm1.0$ & $23.3\pm0.6$ & $22.7\pm4.6$ & $27.7\pm4.9$ & $25.3\pm0.6$ \\
\bottomrule
\end{tabular}%
}
\end{table*}

\subsection{Real-World Deployment}
\label{ssec:real_robot_deployment}
To validate the physical transferability and robustness of the proposed framework, we conducted experiments using a 7-DoF FP3 robotic manipulator. 
We focused on the pick and place task to evaluate the policy's ability to approach, grasp, and manipulate objects.
The system used a zero-shot sim-to-real configuration, where inference was performed by the ensemble policy within the Isaac Lab environment. 
At each control step, the inferred joint position targets were transmitted via a low-latency UDP connection to a custom ROS 2 bridge node. 
This bridge translated simulated targets into ROS 2 messages for a joint position controller, while parallel gripper commands were dispatched to the gripper action server. 
MoveIt 2~\cite{goerner2019} provided the interface to the low-level joint tracking controllers, ensuring a robust execution pipeline. 
Our successful sim-to-real transferred policies and the uncertainty-aware switching logic demonstrates high-fidelity generalisation to physical hardware without the requirement for iterative real-world fine-tuning or domain adaptation.

% To validate the physical transferability and effectiveness of our learned control policies, we conducted qualitative real-world experiments using a physical Franka FP3 robotic arm (illustrated in the supplementary video). 
% We evaluated the system on the pick-and-place task to test the policy’s ability to approach, grasp, and reliably manipulate objects outside of simulation. 
% The control architecture leveraged a zero-shot sim-to-real configuration, where inference was performed using the ensemble policy deployed within the Isaac Lab environment. 
% At each control step, the inferred joint position targets were transmitted via a low-latency UDP connection to a custom ROS 2 bridge node running on the robot's control workstation.
% This bridge processed the incoming data and published the simulated joint targets as ROS2 messages to a joint position controller, while parallel gripper commands were dispatched to the gripper action server. 
% We utilised MoveIt2 to interface with the low-level joint tracking controllers, providing a safe and robust execution pipeline for the physical hardware. 
% During qualitative testing, the physical robot successfully completed the full pick-and-place task in the real world. 
% This successful execution demonstrates that the behaviour cloning policies and the integration of our uncertainty-aware backup controllers, developed entirely within the simulated environment, can be robustly deployed to physical hardware without requiring any supplementary real-world fine-tuning.

%[ADD some method limitations]
\section{Conclusion}

In this work, we presented \textbf{SAFE-CHEM}, an uncertainty-aware imitation learning framework designed to improve the safety and reliability of autonomous robotic manipulation in chemistry laboratories. 
While learning-based methods offer the dexterity required for learning laboratory skills in long-horizon chemistry experiments, their susceptibility to overconfident action prediction poses substantial safety risks. 
To address this, \textbf{SAFE-CHEM} uses an ensemble of BC-RNN policies to quantify epistemic uncertainty online through ensemble action variance. 
By calibrating confidence thresholds using kernel density estimation, our framework introduces an uncertainty-driven switching mechanism that transitions control to a deterministic, rule-based backup controller before a critical failure can occur.
Our empirical evaluations across three representative chemistry manipulation skills (lifting glassware, pick and place of glassware, and high-precision glassware insertion) demonstrated that the proposed hybrid framework significantly improves overall task success rates. 
Crucially, the ensemble of policies reduced the occurrence of catastrophic safety violations, such as dropped glassware or potential chemical spills from tilted beakers, compared to standard single-policy baselines. 
Furthermore, we demonstrated the practical viability of our approach through a successful zero-shot sim-to-real transfer onto a physical FP3 robotic manipulator.
Whilst our method cannot guarantee to eliminate all safety violations, it demonstrably improves task execution. 
Our method is agnostic to policy, task, and environment, and does not require an exhaustive list of all possible failure modes in order to identify and reduce the occurrence of these safety violations. 
Ultimately, \textbf{SAFE-CHEM} bridges the gap between the flexibility of imitation learning and the strict reliability requirements of human-centric laboratory environments. 
Future work will explore extending this uncertainty-quantification framework to incorporate multi-modal sensory inputs, such as force-torque feedback and visual-based policies, with a particular focus on long-horizon multi-skill experiments. 
By equipping autonomous robotic chemists with the introspective ability to assess their own competence, this work provides a critical stepping stone towards the widespread adoption of safer autonomous robotic chemists in laboratories.

\bibliographystyle{IEEEtran}
\bibliography{bibliography}
\end{document}